\documentclass{article}

\usepackage{PRIMEarxiv}

\usepackage{soul}
\usepackage{url}
\usepackage{hyperref}
\usepackage[utf8]{inputenc}
\usepackage{caption}
\usepackage{xcolor}
\usepackage{colortbl}
\usepackage{graphicx}
\usepackage{caption}
\usepackage{courier}
\usepackage{multirow}
\usepackage{subcaption}
\usepackage{color}
\usepackage{todonotes}
\usepackage{amssymb}
\usepackage{gensymb}
\usepackage{enumitem}
\usepackage{booktabs}
\usepackage[linesnumbered, ruled]{algorithm2e}
\usepackage{amsmath,amssymb}
\usepackage[subtle]{savetrees}
\usepackage{algorithmic}
\usepackage{pifont}
\usepackage{xcolor}
\usepackage{etoolbox}
\usepackage{fontawesome5} 
  
\title{Cite Before You Speak: Enhancing Context-Response Grounding in E-commerce Conversational LLM-Agents}

\author{
  Jingying Zeng$^*$, Hui Liu$^*$, Zhenwei Dai$^*$, Xianfeng Tang, Chen Luo,  Samarth Varshney, Zhen Li, Qi He \\
  Amazon \\
  \texttt{zejingyi@amazon.com} \\
}

\begin{document}
\maketitle
\def\thefootnote{*}\footnotetext{Equal Contribution.}

\begin{abstract}

With the advancement of conversational large language models (LLMs), several LLM-based \textbf{C}onversational \textbf{S}hopping  \textbf{A}gents (CSA) have been developed to help customers smooth their online shopping. The primary objective in building an engaging and trustworthy CSA is to ensure the agent's responses about product factoids are accurate and factually grounded. However, two challenges remain. First, LLMs produce hallucinated or unsupported claims. Such inaccuracies risk spreading misinformation and diminishing customer trust. Second, without providing knowledge source attribution in CSA response, customers struggle to verify LLM-generated information. To address both challenges, we present an easily productionized solution that enables a ``citation experience'' to our customers. We build auto-evaluation metrics to holistically evaluate LLM's grounding and attribution capabilities, suggesting that citation generation paradigm substantially improves grounding performance by \textbf{13.83\%}. To deploy this capability at scale, we introduce \textbf{M}ulti-\textbf{U}X-\textbf{I}nference system, which appends source citations to LLM outputs while preserving existing user experience features and supporting scalable inference. Large-scale online A/B tests show that grounded CSA responses improves customer engagement by \textbf{3\%}–\textbf{10\%}, depending on UX variations.

\end{abstract}


\section{Introduction}

\begin{figure*}[htb]
    \vspace{-5pt}
    \centering
    \includegraphics[width=0.9\textwidth]{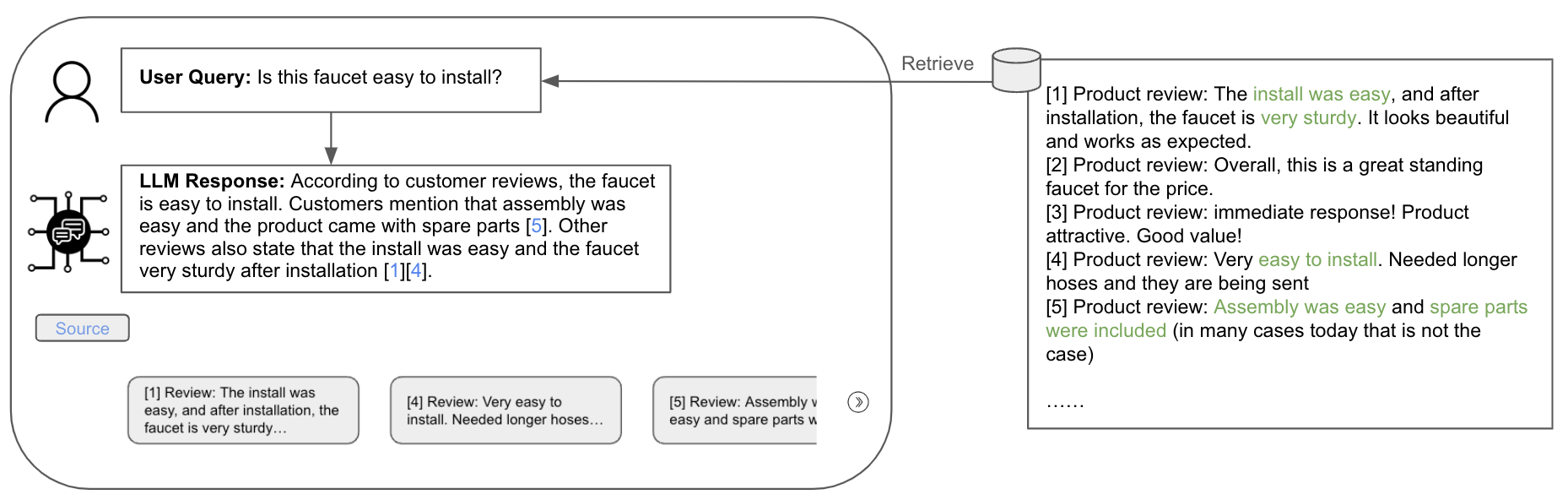}
    \vspace{-5pt}
    \caption{An Overview of Citation Generation Paradigm. For this specific online customer query, Customer Reviews are retrieved as in-context evidences to generate  a more accurate and reliable response with easily verifiable knowledge sources.}
    \label{citation_ux}
    \vspace{-5pt}
\end{figure*}

With the increasing capabilities of LLMs, e-commerce has been revolutionized, shifting from traditional search-and-buy methods to a more interactive and tailored shopping experience. For instance,  Conversational Shopping Agent (CSA) has been introduced in online shopping, aiming at smoothing customers' shopping journey by helping them answer questions about products, identify their shopping intents, and navigate to the most suitable products through conversations. Given the nature of a shopping assistant, one essential ability that the CSA should possess is to provide accurate and relevant answers to product-related questions, where about 66\% of the customer questions are about product facts.

However, accurately answering customer questions with precise product information can be challenging. Firstly, for a given customer query and product, finding correct and reliable product facts is non-trivial. Product descriptions provided by sellers are sometimes inconsistent with actual customer experiences. For example, shoes advertised as ``true to size'' might turn out to be ``oversized'' for the majority of customers. Secondly, even when reliable information is available, LLMs may still hallucinate \cite{ji2023survey, dhuliawala2023chain}. Retrieval-augmented generation (RAG) \cite{shuster2021retrieval,tonmoy2024comprehensive,ayala2024reducing, zeng2024large} can alleviate the hallucination problem in LLMs by augmenting retrieved information directly into the generation process, however, its effectiveness highly depends on the relevancy of the retrieval sources \cite{chen2023understanding}. Our analysis reveals that approximately $3\%$ of claims in CSA-generated responses are not grounded on the product facts provided. When customers verify the sources of the information and discover discrepancies themselves, it will significantly reduce customer-CSA trust, resulting in diminished retention on CSAs over the long run.

In this light, we propose a novel framework to enable ``citation experience'' on CSA that links LLM-generated responses to reliable product facts. In-context Learning (ICL) has been demonstrated as the emerging capability of LLMs \cite{brown2020language,wei2022emergent, dong2022survey, zeng2025examples}. Research has shown the effectiveness of using ICL to instruct LLMs to generate responses while providing citations of the origins in the form of box brackets (i.e. $[n]$) \cite{gao2023enabling}. However, one of the biggest challenges in productionizing the citation generation paradigm as a UX feature on CSAs via ICL is that without explicit model training to enhance the backbone LLM’s attribution capability, the quality of the generated citations might be compromised and may interfere with other UX features that have been trained and deployed. To ensure both quality and efficient latency to build a Minimum Viable Product (MVP), we propose a solution utilizing ICL accompanied by \underline{M}ulti-\underline{U}X \underline{I}nference (MUI) technique to address the aforementioned issues at decoding time. 
Our approach comprises three key components: 
\begin{itemize}
    \item \textbf{Easily Productionized Solution for Knowledge Source Attribution in LLM-Generated Responses}: Inspired by LLM-based search systems like Perplexity or Gemini, we implement an easily productionizing solution via ICL to enable LLM citing information sources directly in LLM responses. By appending citations to statements, customers can easily trace information back to verified product reviews and descriptions. We also address inference challenges through MUI with inference optimization techniques.
    \item \textbf{Leveraging multi-perspective product information}:  we dynamically supply the generative LLMs with multi-perspective in-context content, such as merchants’ product descriptions, customer reviews, customer Q\&A, and etc., to let the CSA generate more well-rounded, reliable answers. The multi-perspective knowledge source provided as in-context can mitigate inaccuracies stemming from unreliable seller-provided marketing descriptions.
    \item \textbf{Auto-Metrics and Benchmarks for Evaluation and Accelerating Iteration}: We develop two sets of novel metrics with scalable benchmarks to holistically assess the grounding and attribution capabilities of LLMs, as well as diagnosis metrics to investigate LLM's intrinsic deficiency on context utilization. 
\end{itemize}
Our experiments demonstrate that incorporating citation generation paradigm in LLM's response generation can increase response grounding by \textbf{13.83\%}, which significantly improves response credibility on product-related information seeking questions. Our insights also reveal that the citation generation paradigm can substantially reduce hallucinations by providing ``refusal signals'' (e.g. ``The reviews do not provide information about ...'') when sufficient product information cannot be obtained from retrieval sources. 

In this work, our key contributions are: \ding{202} Propose a lightweight citation-generation solution, paired with an automated, holistic evaluation framework—both engineered for rapid MVP deployment to gather real-world customer feedback. \ding{203} Introduce Multi-UX Inference (MUI), a technique that prevents ICL-based UX features from degrading one another’s performance; and an efficient request orchestration layer that scales LLM inference \ding{204} Demonstrated impacts of citation generation paradigm on improving response grounding by \textbf{13.83\%} and customer engagement rate by \textbf{3-10\%}.


\section{Citation Generation}

In this section, we will describe the citation generation paradigm in more detail. With this paradigm, we aim to: (1) improve CSA's response grounding and alleviate hallucinations; (2) enhance customer-CSA trust by showing that CSA responses are grounded in real product facts, which is accessible and easily verifiable for customers.


Figure~\ref{citation_ux} shows an example of the citation generation task and UX design. Given a query, a retriever retrieves and ranks the related product information and real customer reviews, then feed into LLM to generate a response with the ``citation'' marks in enclosed box brackets, indicating the knowledge source in the sentence. The UX displays the generated response, with the source widgets provided at the bottom, where each piece is an evidence source (i.e., customer reviews, product information). By clicking the widget, the customer is directed to the evidence page through the evidence identifier.

\subsection{Problem Formulation}

In retrieval-augmented generation, for a given online customer query $q$, the retriever first retrieves a set of evidences $\mathcal{E}$. A generative LLM, $\mathcal{M}$ takes the prompt, which consists an online customer query $q$, the retrieved evidence set $\mathcal{E}$, and a system prompt $P$ to produce a response $\mathcal{R}=\mathcal{M}(q, \mathbb{E}, p)$. 
The response $\mathcal{R} = \{s_1, s_2, ..., s_n\}$ consists of $n$ sentences. Each sentence $s$ cites from a list of evidences $\mathcal{E}^s = \{e_1^s, ... , e_l^s\} \subseteq \mathcal{E}$. If a sentence is generated without any citations, $\mathcal{E}^s = \emptyset$. 


\subsection{Citation Generation Methodology}
With the increasing capabilities of large language models (LLMs), in-context learning (ICL) has been demonstrated as an emerging ability of LLMs to follow instructions and learn from a few examples in the in-context~\cite{dong2022survey}. In-context learning is a powerful and convenient approach for adapting LLMs for downstream tasks. Compared with other approaches, ICL provides an efficient way to build Minimum Viable Products (MVPs) and test user feedback. Post-hoc approach~\cite{ye2024effective} adopts Natural Language Inference (NLI) models to capture the entailment relations between each evidence and each sentence in CSA's response, and add the evidence citations to the response if the evidence supports the sentence. However, adding citations retroactively can significantly increases the latency of the response generation. Fine-tuning approach \cite{ye2024effective, huang2024learning, zhang2024longcite}, on the other hand, can further adapt the pre-trained LLMs to generate responses with attributable sources. However, it requires a holistic training approach to integrate attributable information generation into the answer generation process without degrading the model's other fundamental capabilities. Additionally, model training requires high-quality data to be collected and annotated, which is time-consuming and labor-intensive. With all these considerations, we employ ICL-based approach \cite{gao2023enabling} to enable citation generation and dynamically provide LLM multi-perspective sources—product descriptions, customer reviews, and Q\&A as in-context to increase response credibility.

\section{Automatic Evaluation}
Next, we will describe the automatic evaluation pipeline used to \textbf{accelerate iteration} and \textbf{assess citation generation quality}, which includes both automatic metrics and scalable benchmarks for a comprehensive assessment. This section focuses on the auto-metrics, while we leave the details of benchmark creation in Section~\ref{Experiment_result}.

The evaluation framework utilizes LLM-as-a-judge \cite{zheng2024judging} and Natural Language Inference (NLI) model $\Phi$ to automate the process. Specifically, an NLI model checks whether a text $t$ (e.g. a sentence $s$ or a claim $c$) can be supported by an evidence $e$. A text $t$ is supported by the evidence $e$ only when $e$(premise) entails $t$ (hypothesis), denoted as $\Phi(e,t)=1$. On the other hand, $t$ cannot be supported by $e$ if $\Phi(e,t)=0$. For simplicity, we prompt an LLM through instructions with few-shot examples to perform NLI task in this work. With prompt optimization, we empirically observe high correspondence between human and LLM annotations.

\subsection{Evaluate LLM Grounding Ability}

The key benefit of the citation generation paradigm is to improve the LLM grounding ability and generate reliable answers to customers. To answer the questions about the product facts, ACA will retrieve the product information such as product descriptions, reviews, and generate response based on the retrieved information. Therefore, we develop the metric \textbf{Claim Grounding Rate (CGR)} to quantify how many claims in the response can be grounded by retrieved evidences. For a given generated response $\mathcal{R}=\mathcal{M}(q,\mathcal{E},p)$, we first utilize an LLM to decompose the response $\mathcal{R}$ into a set of claims 
    $\mathcal{C}=\{c_1,...,c_m\}$ where $m$ is the total number of claims generated. 
For each claim $c_j$, we check whether the claim can be supported by any evidence $e$ from the retrieved evidence set $\mathcal{E}$. We then quantify the LLM's grounding capability using the percentage of generated claims that can be grounded on the evidence set $\mathcal{E}$, which is the Claim Grounding Rate defined as:
\begin{equation}
   CGR = \frac{m^{\text{ground}}}{m},
\end{equation}
where $m^{\text{ground}}$ is the total number of claims that can be supported by the evidence from $\mathcal{E}$. A low CGR means most of the generated claims cannot be grounded by the retrieved evidences, indicating that either the model suffers from severe hallucinations or it cannot leverage the retrieved evidence to generate the response.

\subsection{Evaluate LLM Attribution Ability}
The intrinsic attribution capability of LLMs determines how well the responses are grounded and the quality of the generated citations as well. We define the following metrics to evaluate the LLM's attribution capability and the generated citation quality.

\begin{itemize}
    \item \textbf{Correct Citation Rate (CCR)}: CCR evaluates how many cited evidences are entailed by the corresponding claims. Given a sentence $s$ and its associated cited evidences $\mathcal{E}^s = \{e_1^s, ... , e_l^s\}$, we say the citation $e^{s}_i$ is correct if the sentence $s$ is entailed by $e_i^s$ through an NLI model (i.e. $\Phi(e^{s}_i, s)=1$). Assume the LLM response $\mathcal{R}$ cited $r$ evidences in total, where $r = \sum_{s\in R} |\mathcal{E}^s|$, and $r^\text{entail}$ is the number of cited evidences that are entailed. We define the Correct Citation Rate:
    \begin{equation}
        CCR = \frac{r^\text{entail}}{r}
    \end{equation}

    \item \textbf{Perfect Sentence Rate (PSR)}: PSR measures percentage of sentences that correctly cite the sources. Assume the LLM response $R$ contains $n^{\text{cited}}$ sentences with citations. We define the sentence $s$ is ``perfect'' if all the cited evidences of the sentence are correct, i.e. $\Phi(e^{s}_i, s)=1$ for all $e^{s}_i \in \mathcal{E}^s$. Denote $n^{\text{p-cited}}$ as the number of ``perfect'' sentences. We can naturally define Perfect Sentence Rate (PSR) as:
    \begin{equation}
        PSR = \frac{n^{p-cited}}{n^{\text{cited}}}
    \end{equation}
    
    \item \textbf{Sentence with Citation Rate (SCR)}: SCR computes the percentage of sentences having citations. Assume among $n$ sentences generated, there are $n^\text{cited}$ sentences that contain the generated citations, then we define SCR as:
    \begin{equation}
        SCR = \frac{n^{cited}}{n}
    \end{equation}

\end{itemize}

\subsection{Evaluating LLM's Evidence Utilization}
LLMs are well-known for the ``lost-in-the-middle'' challenge, where it is hard to fully utilize the in-context information \cite{liu2024lost}. To diagnose the intrinsic deficiency of LLM used in RAG system, we define \textbf{Evidence Utilization Rate (EUR)} to quantify how many evidences are utilized by the LLM. Specifically, let $k^{\text{ground}}$ be the number of evidences cited in the LLM response, we define the Evidence Utilization Rate (EUR) as
    \begin{equation}
    EUR = \frac{k^{\text{ground}}}{|\mathcal{E}|} \cdot (1 - \frac{|\mathcal{E}| - k^{\text{ground}}}{|\mathcal{E}|^2})    
    \end{equation}
Note that EUR penalizes smaller number of retrieved evidences $|\mathcal{E}|$. For example, $5$ out of $10$ evidences cited in the LLM response will be better than citing $1$ out of $2$ evidences. A low EUR means that either the model does not adequately utilize the evidences or the retrieved evidences are not closely related to the query.


\section{Offline Experiment Setups and Evaluation Results} \label{Experiment_result}
We leverage both product information and customer reviews as retrieved evidences, use an internal LLM as the backbone model in the CSA to conduct the experiments, and employ Llama-3.1-70B as the judge to obtain the results. In this section, we will present the research questions that we investigated, how to create benchmarks and conduct experiments, and the insights that we discovered.

\textbf{Datasets:} We leverage CSA's data and retrieved product information to automate the creation of the two benchmarks. One benchmark simulates the \textbf{ideal scenario} where \textbf{all} the retrieved evidences are relevant to the question. The synthetic benchmark is created using the query ``\textit{What do customers say about the pros and cons?}'' for a wide range of different types of products, which ideally should yield mostly relevant customer reviews. Using this synthetic benchmark, we can decouple the effects of noisy retrieval and have a true understanding on the model's capabilities. The other benchmark simulates the \textbf{real-world scenario} where retrieved evidences contain noisy information. The real-world benchmark is created by sampling from the CSA's conversational data, where only 41.1\% retrieved evidences are relevant to the customer queries.

\textbf{Baselines:} 
We evaluate the grounding and attribution capabilities based on the following three prompts: 
\begin{itemize}
    \item \textbf{Vanilla Prompt}: the prompt which directly ask LLM to answer question given the retrieval evidence as context.
    \item \textbf{Guided Prompt}: the prompt optimized  via ICL instructions and few-shot examples (w/o citation instructions) to help LLM answer shopping related context and requirements.
    \item  \textbf{Citation Prompt}: the prompt with citation instructions to instruct LLM generate responses while citing the sources of the generated statements.
\end{itemize}


\begin{table}[h!]
\centering
\caption{Evaluate Grounding \& Evidence Utilization with Synthetic Dataset}\label{prompt_optimization}
\begin{tabular}{l|ccc}
\hline
\multicolumn{1}{c}{} & Vanilla Prompt & Guided Prompt & \multicolumn{1}{l}{Citation Prompt} \\
\hline
\cellcolor[HTML]{E6E6E6}CGR                  & 93.91\%        & 96.52\%       & \textbf{98.65\%}                    \\
\cellcolor[HTML]{E6E6E6}EUR                  & 72.50\%        & 78.99\%       & \textbf{82.17\%}   \\
\hline
\end{tabular}
\end{table}

\textbf{RQ 1: With ICL, can we improve the LLM's grounding and evidence utilization capabilities, and can we further enhance these capabilities by instructing the LLM to attribute its knowledge source?} We first use the synthetic benchmark to evaluate the true grounding and evidence utilization capabilities of LLM. Table~\ref{prompt_optimization} shows that with some proper prompt optimizations, we can improve the response grounding and evidence utilization comparing to the Vanilla Prompt. However, without instructing LLMs to cite the sources of the generated statements, there are still over 3.5\% claims in the responses that cannot be supported by the evidence. If customers capture these incorrect statements, they may lose trust in the entire response, even if the CSA provides mostly correct answers. We hypothesize that the grounding and evidence utilization capabilities of the LLM can be elicited by instructing it to generate responses with attributed sources. According to Table~\ref{CGR_result}, our experimental results verify this hypothesis, showing a 2.21\% improvement in response grounding on synthetic benchmark. When applying the citation generation paradigm to a real-world benchmark, we observed a much more significant improvement in response grounding, where the CGR improved by \textbf{13.83\%}.

\begin{table}[h!]
\centering
\caption{The CGR of Guided Prompt v.s Citation Prompt}\label{CGR_result}
\begin{tabular}{l|ll}
\hline
                    & \cellcolor[HTML]{E6E6E6}Guided Prompt & \cellcolor[HTML]{E6E6E6}Citation Prompt \\
                    \hline
Synthetic Benchmark     & 96.52\%         & \textbf{98.65\%}         \\
Real-world Benchmark    & 83.86\%         & \textbf{95.46\%}      \\
\hline
\end{tabular}
\end{table}



\textbf{RQ 2: When no sufficient product information retrieved for a given query, can we use citation generation paradigm to prevent LLM from hallucination?} When none of the retrieved evidences are useful in answering the customer questions about the given product, we expect LLM to generate ``refusal signals'' rather than hallucinate or rely on its parametric knowledge in answer generation. For instance, when a customer asks ``Are there reviews from customers who've had them for over a year?'' and none of the retrieved real customer reviews contain information about long-term durability beyond a year for this given product, an ideal response for this query would be providing some ``refusal signals'' such as ``I don't have enough information about ...''. To analyze LLM's performance on this scenario, we sample $60$ queries from CSA's conversational data, where none of the retrieved evidences are relevant to the corresponding questions. Based on our evaluation, \textbf{all} the responses generated using the citation prompt provided ``refusal signals'', while only 36.67\% of the responses generated from the Guided Prompt perfectly match the product facts, indicating that the citation-based approach significantly enhances the reliability of the generated responses.

\textbf{RQ 3: How is the performance on enabling citation generation via ICL?}
As can be seen from Table~\ref{attribution_result}, without explicit model training, LLM is able to correctly attribute its knowledge source above $70\%$ (CCR) of the time. However, only about $50\%$ (PSR) of the sentences generated in the response are perfectly attributed. This suggests that, with ICL, there remains a performance ceiling for generating fully cited responses without additional training.

\begin{table}[h!]
\centering
\caption{Evaluate Attribution Capability of Citation Prompt} \label{attribution_result}
\begin{tabular}{l|ccc}
\hline
                    & \cellcolor[HTML]{E6E6E6}CCR  & \cellcolor[HTML]{E6E6E6}PSR & \cellcolor[HTML]{E6E6E6}SCR \\
 \hline                   
Synthetic Benchmark    & 72.52\%                      & 52.94\%    & 86.90\%                                           \\
Real-world Benchmark & 71.13\%                      & 50.07\%    & 52.48\%                \\
\hline
\end{tabular}
\end{table}

\textbf{RQ 4: Does the number of retrieved evidences have an impact on citation generation?} We also studied the impact of number of evidences on  citation generation on real-world benchmark in Table~\ref{num_evidences}.
By varying the number of evidences from 24 to 5, the three metrics do not show significant change, though we observe slightly increased CCR and EUR, and decreased SCR.
Such results align with existing research that fewer evidences may reduce the LLMs' confusion in the context, which could hence increase the correction citation rate and enhance evidence utilization. 
However, fewer evidences may reduce the information that is available to LLMs when generating responses, which will reduce the percentage of sentences with citations consequently.


\begin{table}[h!]
\centering
\caption{Ablation Study on the Number of Evidences}\label{num_evidences}
\begin{tabular}{cccc}
\hline
\rowcolor[HTML]{E6E6E6} 
\hline
\# of evidences & CCR     & SCR     & EUR     \\
\hline
24              & 71.13\% & 52.48\% & 54.31\% \\
5               & 74.77\% & 48.80\% & 58.67\% \\
\hline
\end{tabular}
\end{table}

\section{Online Deployment and Evaluation}

\textbf{Deployment considerations:} The CSA system is deployed in multiple UXs while some UXs such as recommending products do not need citations. The existing product generates the responses to all the UXs using a unified prompt to guarantee the consistency between different UXs. To enable citation experience in production, the most straightforward approach is directly add citation instructions to the unified prompt. However, our analysis shows that this approach leads to performance regressions in other UXs, and LLM could hallucinate to add citations to the UXs that do not need citations.
To maintain the answer quality and avoid multiple LLM inference calls for the same customer request, we propose \textit{Multi-UX-Inference} that leverages paged attention~\cite{kwon2023efficient} to cache the shared prompt. Specifically, when introducing a new UX via ICL, the corresponding instructions are appended to the end of the original prompt such that the KV cache of the original prompt can be reused. During the prefilling stage, we compute the key-value (KV) cache for the entire prompt. The KV cache corresponding to the citation instructions is stored on separate pages. In the decoding stage, we use the full prompt to generate responses with citations, while excluding the citation instructions page when generating responses for other UXs. We found this approach could maintain comparable answer quality without multiple prefilling, while boosts the inference speed and saves the GPU memory cost.

\begin{figure}[htb]
    \vspace{-5pt}
    \centering
    \includegraphics[width=0.8\textwidth]{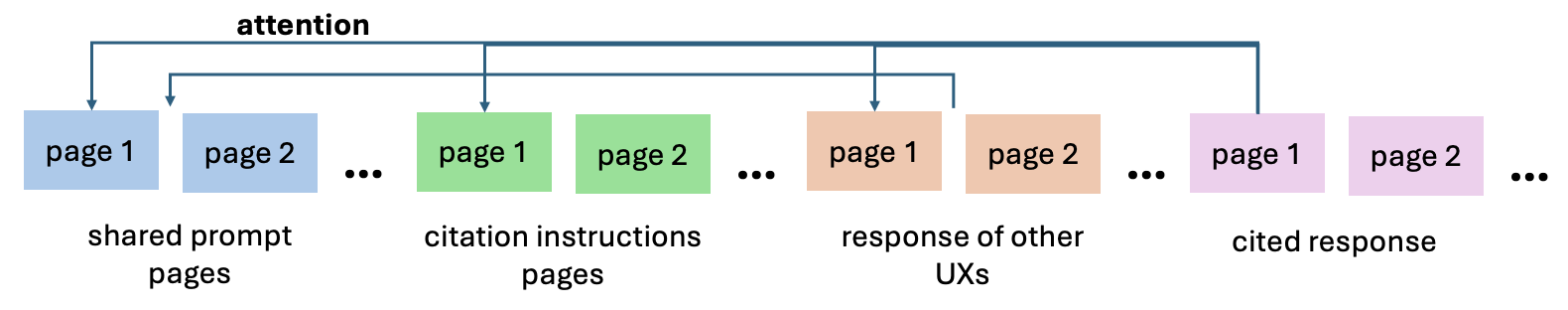}
    \vspace{-5pt}
    \caption{Multi-UX-inference design with paged-attention.}
    \label{citation_ux}
    \vspace{-5pt}
\end{figure}

\textbf{Online A/B test results}: 
We ran a 2-week online A/B test to evaluate the impact of grounded CSA answers, where the customers are evenly partitioned into control and treatment groups.  The large-scale A/B test involves over 10 million customers and shows that improving the system's answer grounding capabilities can increase customer engagement rate of CSA by \textbf{3–10\%} on different UXs.


\section{Conclusion}
In this paper, we present a training-free and easily productionized solution that enables CSA to generate responses and citations of the original knowledge sources in one pass. Our solution introduces MUI to address the deployment challenge, ensuring that introducing new ``citation experience'' does not compromise the performance of existing interfaces. Our experiments highlight the effectiveness of utilizing citation generation paradigm to reduce LLM's hallucination during answer generation. We hope our research can foster the development of trustworthy CSAs in the future, providing customers a better online shopping experience.

\bibliographystyle{unsrt}  
\bibliography{references}

\end{document}